\documentclass{svproc}
\usepackage{url}

\usepackage{booktabs}
\usepackage{makecell}
\usepackage{breakcites}
\usepackage{silence}
\usepackage{caption}
\usepackage{multirow}
\usepackage{graphicx}
\usepackage{amsmath,amssymb}
\usepackage{xcolor}
\usepackage{siunitx}
\usepackage{float}
\usepackage{placeins}
\usepackage{etoolbox}
\usepackage{tcolorbox}
\tcbuselibrary{listings, breakable, skins}
\pretocmd{\subsection}{\FloatBarrier}{}{}
\pretocmd{\subsubsection}{\FloatBarrier}{}{}
\renewcommand{\arraystretch}{1}

\lstdefinelanguage{json}{
    basicstyle=\ttfamily\scriptsize, 
    numbers=none,
    showstringspaces=false,
    breaklines=true,
    frame=single,
    framerule=1pt,
    tabsize=2,
    morestring=[b]
}

\begin{document}
\mainmatter              
\title{TERAG: Token-Efficient Graph-Based Retrieval-Augmented Generation}
\titlerunning{TERAG}  
%
\author{Qiao Xiao\inst{1} \and Hong Ting Tsang\inst{2} \and Jiaxin Bai\inst{2}}
\authorrunning{Qiao Xiao et al.}
\tocauthor{Qiao Xiao, Hong Ting Tsang, and Jiaxin Bai}
\institute{Cornell University, United States\\
\email{qx226@cornell.edu}
\and
Hong Kong University of Science and Technology, Hong Kong\\
\email{httsangaj@connect.ust.hk}, \email{jbai@connect.ust.hk}}

\maketitle              

\begin{abstract}
Graph-based Retrieval-augmented generation (RAG) has become a widely studied approach for improving the reasoning, accuracy, and factuality of Large Language Models (LLMs). However, many existing graph-based RAG systems overlook the high cost associated with LLM token usage during graph construction, hindering large-scale adoption. To address this, we propose \textbf{TERAG}, a simple yet effective framework designed to build informative graphs at a significantly lower cost. Inspired by HippoRAG, we incorporate Personalized PageRank (PPR) during the retrieval phase, and we achieve at least 80\% of the accuracy of widely used graph-based RAG methods while consuming only 3\%-11\% of the output tokens. With its low token footprint and efficient construction pipeline, TERAG is well-suited for large-scale and cost-sensitive deployment scenarios.
\end{abstract}


\section{Introduction}
Retrieval-Augmented Generation has emerged as an important framework to mitigate hallucinations and ground Large Language Models in external knowledge, enhancing their reliability in specialized domains like medicine~\cite{liu2024surveymedicallargelanguage}, law~\cite{fan2024goldcoingroundinglargelanguage}, and education~\cite{ghimire2024generativeaieducationstudy}. While traditional RAG retrieves unstructured text snippets, recent advancements have shifted towards graph-based RAG, which leverages knowledge graphs (KGs) to model structured relationships between information entities~\cite{peng2024graphretrievalaugmentedgenerationsurvey}. This structured approach enables more accurate multi-hop reasoning and provides greater transparency into the model's decision-making process~\cite{Edge2024FromLT}.

However, the superior performance of state-of-the-art graph-based RAG systems, such as AutoSchemaKG~\cite{bai2025autoschemakgautonomousknowledgegraph} and Microsoft's GraphRAG~\cite{Edge2024FromLT}, comes at a staggering cost. These methods rely heavily on extensive LLM calls for node extraction, relationship definition, and schema induction, resulting in extremely high token consumption. This dependency makes graph construction prohibitively expensive; for instance, indexing a mere 5GB of legal documents was recently estimated to cost as much as \$33,000~\cite{huang2025ketragcostefficientmultigranularindexing}. In practice, such a financial burden poses a major barrier to scalable deployment, making cost-effectiveness as crucial a metric as accuracy.

\begin{figure}[t]
    \centering
    \caption{Overall pipeline of the proposed TERAG framework. The process consists of lightweight concept extraction with LLMs, followed by efficient non-LLM clustering and graph construction.}
    \label{fig:pipeline}
    \includegraphics[width=0.8\linewidth]{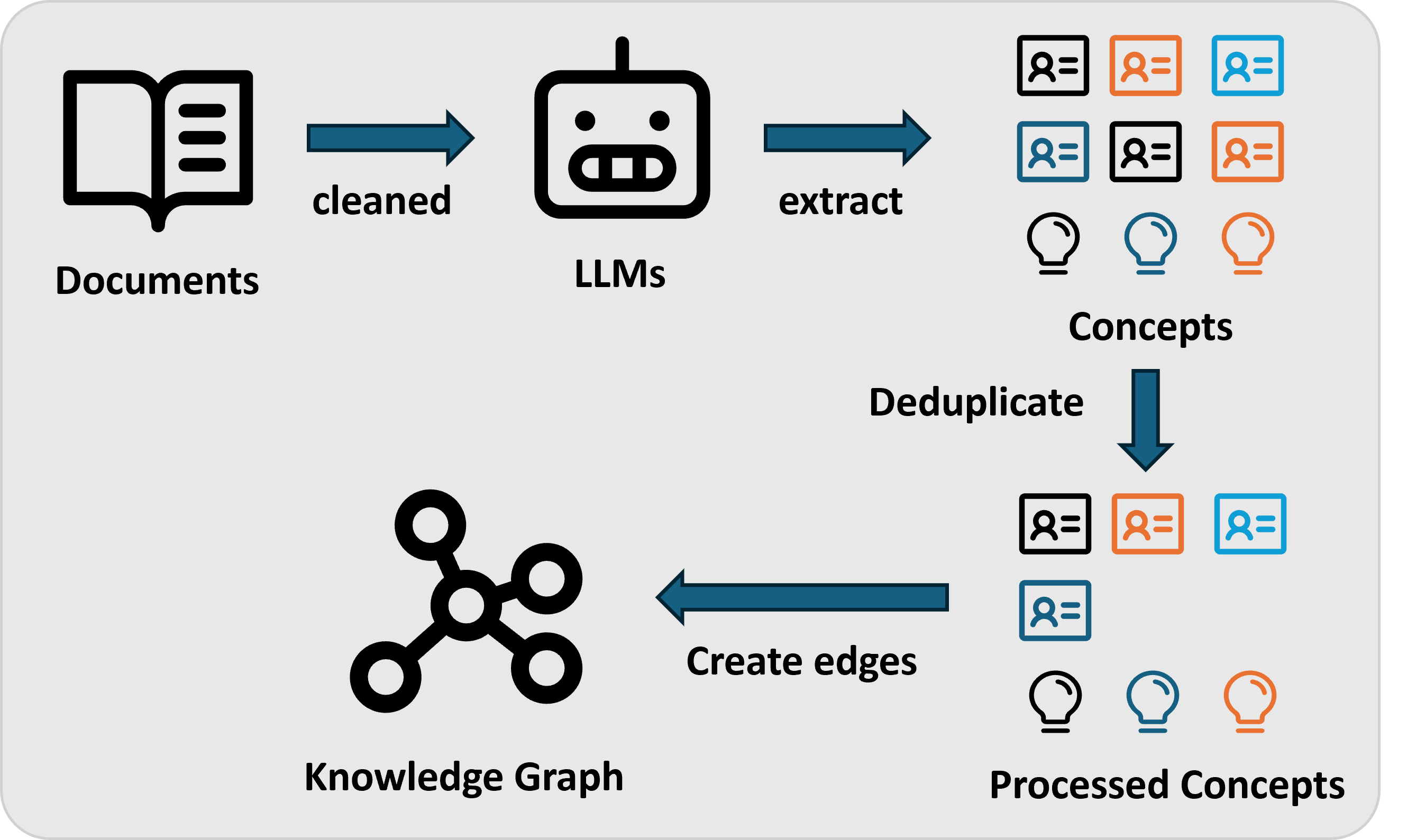}
\end{figure}

\begin{figure}[ht]
    \centering
    \caption{Graph structure of TERAG. The figure illustrates how lightweight concept extraction with LLMs,
    followed by non-LLM clustering and graph construction, leads to an efficient knowledge graph.}
    \includegraphics[width=0.8\linewidth]{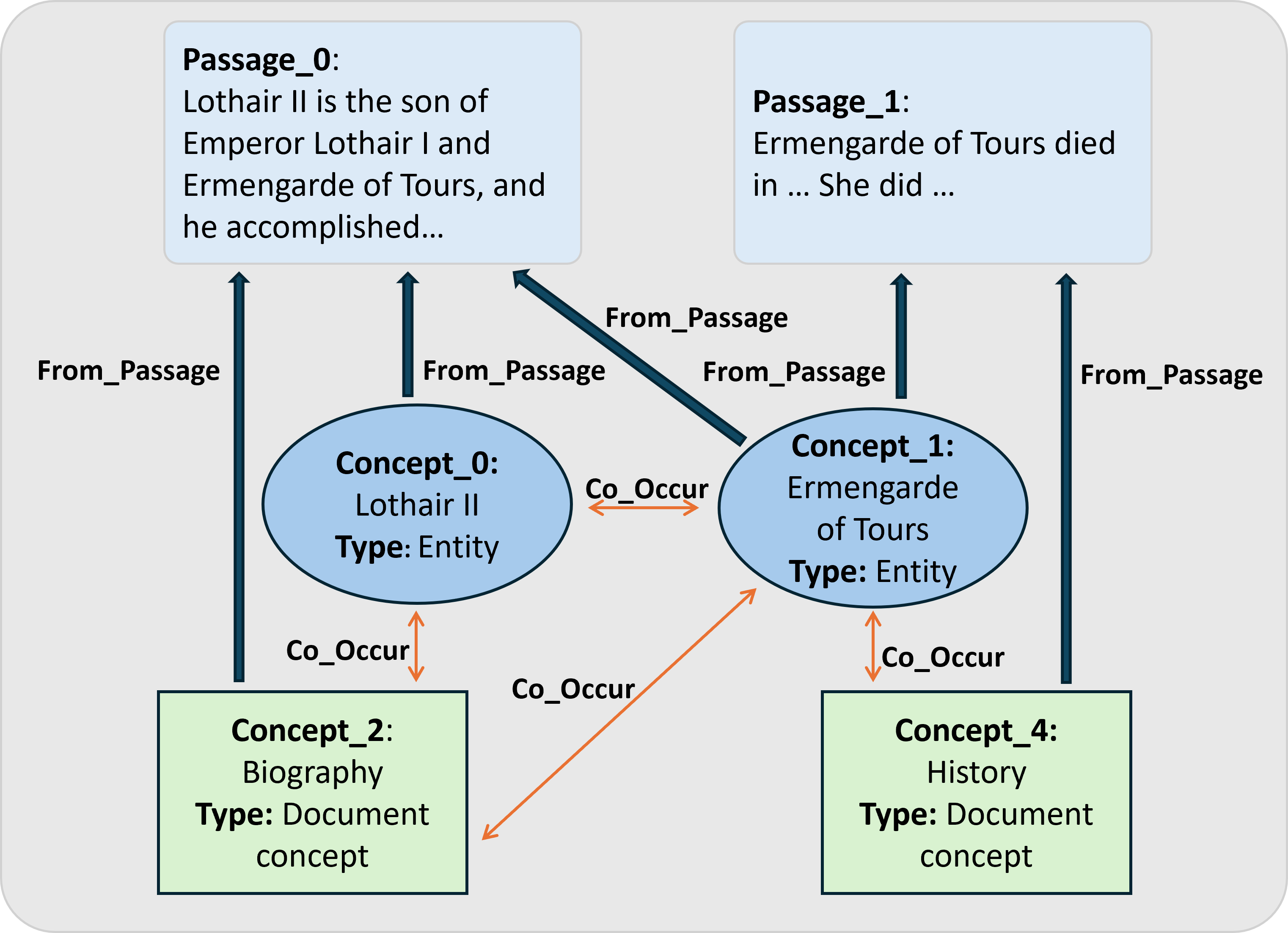}    
    \label{fig:graph_structure}
\end{figure}

To address this critical trade-off between cost and performance, we propose TERAG, a lightweight framework that minimizes LLM usage while leveraging their strengths in concept extraction. The overall pipeline is shown in Figure~\ref{fig:pipeline}. Instead of relying on multiple rounds of expensive LLM reasoning for graph construction, our method uses a few carefully designed prompts to extract multi-level concepts. These concepts are then structured into an effective knowledge graph using efficient, non-LLM methods, as illustrated in Figure~\ref{fig:graph_structure} This process is illustrated with a real case from the 2WikiMultihopQA dataset. The two passages, while topically related, lack a direct connection that simple semantic retrieval could exploit. By extracting concepts from each passage and linking them with a \texttt{co\_occurrence} edge, TERAG successfully connects them via key semantic information. This enables a successful retrieval for the question, ``What is the death date of Lothair II's mother? ``---a query that would likely fail with retrieval methods based only on direct semantic similarity retrieval.

For the retrieval phase, inspired by HippoRAG~\cite{gutierrez2024hipporag,gutiérrez2025ragmemorynonparametriccontinual}, we apply Personalized PageRank to the constructed graph. This approach enhances retrieval effectiveness without requiring additional LLM calls. By focusing LLM usage on initial, lightweight tasks, TERAG strikes a favorable balance between efficiency and effectiveness.

Our main contributions are as follows:
\begin{itemize}
    \item We propose TERAG, a simple yet effective graph-based RAG framework designed specifically to minimize LLM token consumption during graph construction.
    \item We demonstrate that TERAG reduces output token usage by 89-97\% compared to other widely-used graph-based RAG methods, offering a highly cost-effective solution.
    \item We show that despite its lightweight design, TERAG achieves competitive retrieval accuracy on three standard multi-hop question-answering benchmarks.
\end{itemize}
The remainder of this paper is organized as follows: Section 2 formalizes the problem, Section 3 discusses related work, Section 4 details our proposed pipeline, Section 5 presents experimental results, and Section 6 provides an ablation study.

\section{Related Work}
This section traces the development of RAG, from foundational ``naive`` methods to more powerful Graph-based systems. We conclude by analyzing the critical challenge of token efficiency and the resulting cost-performance trade-off that motivates our work.

\subsubsection{Naive RAG}
While pretrained language models have demonstrated a remarkable ability to internalize knowledge from training data, they possess significant limitations. Notably, their parametric memory is difficult to update, and they are prone to generating factually incorrect information, a phenomenon known as "hallucination"~\cite{petroni-etal-2019-language,marcus2020decadeaistepsrobust}. To mitigate these issues, Retrieval-Augmented Generation was introduced to ground model responses in an external, non-parametric knowledge source \cite{lewis2021retrievalaugmentedgenerationknowledgeintensivenlp}.
Initial approaches, often termed ``naive RAG,`` retrieve isolated text chunks based on vector similarity to a user's query. Although this method enhances factual accuracy, it fundamentally overlooks the interconnected nature of information. Consequently, it struggles with complex, multi-hop questions that require synthesizing insights from multiple, related documents \cite{hu-etal-2025-grag}. This core limitation paved the way for more advanced methods like Graph-based RAG, which explicitly models the relationships between knowledge entities to enable more sophisticated reasoning.

\subsubsection{Graph RAG}
The emergence of Graph-based RAG addresses the primary limitation of its predecessors: the failure to model complex relationships between text chunks. Graph RAG leverages a pre-constructed knowledge graph, utilizing structural elements like nodes and paths to enrich the retrieval process and access more interconnected information~\cite{peng2024graphretrievalaugmentedgenerationsurvey}. This paradigm has inspired a wave of powerful systems, including Microsoft's GraphRAG~\cite{Edge2024FromLT}, HippoRAG~\cite{gutierrez2024hipporag}, AutoSchemaKG~\cite{bai2025autoschemakgautonomousknowledgegraph} and ThinkOnGraph~\cite{sun2024thinkongraphdeepresponsiblereasoning}, many of which have demonstrated impressive performance and garnered significant industry attention, including Microsoft, NebulaGraph, and AntGroup~\cite{graphrag_github_misc_2025,NebulaGraphGraphRAG,xue2023dbgpt}. 

The strength of these systems stems from a sophisticated and often costly indexing pipeline, 
where frameworks like GraphRAG~\cite{Edge2024FromLT}, LightRAG~\cite{guo2025lightragsimplefastretrievalaugmented}, and MiniRAG~\cite{fan2025miniragextremelysimpleretrievalaugmented} 
rely on Large Language Models (LLMs) to construct a knowledge graph from raw text. 
This process typically involves extracting entities and their connections as structured 
(subject, relation, object) triplets, and in some cases, generating node summaries or community reports~\cite{Edge2024FromLT}. 
While powerful, this deep integration of LLMs during indexing lead to substantial token consumption, 
creating a significant cost barrier for large-scale, real-world adoption.

\subsubsection{The Challenge in Graph RAG}
In response to prohibitive costs, research has diverged into two main efficiency-focused directions. The first, exemplified by systems like LightRAG and MiniRAG~\cite{guo2025lightragsimplefastretrievalaugmented,fan2025miniragextremelysimpleretrievalaugmented}, prioritizes creating structurally lightweight graphs, though this can paradoxically increase indexing time and token consumption. A second path, therefore, which includes frameworks like LazygraphRAG~\cite{edge2024lazygraphrag}, KET-RAG~\cite{huang2025ketragcostefficientmultigranularindexing}, and our work, TERAG, concentrates directly on minimizing the cost of the indexing process itself. Within this approach, while KET-RAG aims to reduce the overall cost from indexing to retrieval, our work with TERAG focuses aggressively on optimizing output tokens during the construction phase. We prioritize minimizing output token as it is the most critical factor during graph construction.

\section{Problem Definition}

In this section we formalize our setting by defining token consumption, the structure of the graph, and the optimization objective.

\subsection{Token Consumption}
In graph-based Retrieval-Augmented Generation (RAG), the \textbf{graph construction phase} is a primary driver of token consumption. This process can be so resource-intensive that Microsoft's official documentation explicitly advises users to start with small datasets to manage costs and processing time~\cite{graphrag_github_misc_2025}. Therefore, a thorough analysis of token usage is essential for evaluating the overall computational overhead and efficiency of these systems. For the purposes of this paper, we categorize tokens into three distinct types:

\begin{itemize}
    \item \textbf{Input Tokens}: These are the tokens provided to the LLMs as context. This includes the content of the document or passage being processed, as well as any system prompts and task-specific instructions.
    
    \item \textbf{Output Tokens}: Also known as completion tokens, these are the tokens generated by the LLM during the autoregressive decoding process. The generation of these tokens is typically more computationally intensive per token compared to processing input tokens, as they must be produced sequentially, one after another~\cite{zhou2024surveyefficientinferencelarge}.
    
    \item \textbf{Total Tokens}: This represents the sum of input and output tokens, formally expressed as $T_{\text{total}} = T_{\text{in}} + T_{\text{out}}$.
\end{itemize}

Through comparing the token consumption of different RAG methods, particularly the total tokens and output tokens, we obtain a practical evaluation metric for assessing the efficiency of graph construction and retrieval. 

\subsection{Graph Definition}
To address the high token consumption outlined above, our framework builds a knowledge graph specifically designed for efficiency. Unlike prior works that often construct based on triple extraction and multi-relational between entities, we adopt a simpler, more streamlined structure. This graph is composed of only essential semantic units which prove sufficient for robust reasoning while drastically reducing the cost of construction.

\subsubsection{Graph Type}
We use a directed, unweighted graph.
We represent the graph as $G=(V,E)$, where $E \subseteq V \times V$ denotes directed edges.
Since our graph is unweighted, we simply record the existence of an edge $(u,v)$ .
In practice, we store $E$ as adjacency lists for efficient neighborhood expansion.

\subsubsection{Node Types}
The node set is
\[
V = V_{\mathrm{pas}} \;\cup\; V_{\mathrm{con}},
\]
where $V_{\mathrm{pas}}$ are \emph{passage} nodes and $V_{\mathrm{con}}$ are \emph{concept} nodes. 
Concept nodes include both named entities and broader document-level concepts extracted by LLM. 
We apply normalization and remove duplicates to merge repeated concepts before edge construction.

\subsubsection{Edge types}
We define three types of edges between nodes:

\begin{description}
  \item[From-passage edges ($E_{\mathrm{pa}}$)] For each concept node $u \in V_{\mathrm{con}}$ and its supporting passage $p \in V_{\mathrm{pas}}$, we add a directed edge $(u \to p)$ labeled \texttt{has\_passage}, preserving provenance information.

  \item[Co-Occurrence edges ($E_{\mathrm{co}}$)] If two concept nodes $u,v \in V_{\mathrm{con}}$ appear in the same passage, we add bidirectional edges $(u \to v)$ and $(v \to u)$ labeled \texttt{co\_occurrence}, avoiding duplicates to reduce graph density.
\end{description}

The complete edge set is therefore
\[
E = E_{\mathrm{pa}} \cup E_{\mathrm{co}},
\]
with co-occurrence and cluster edges treated as single bidirectional pairs to improve efficiency in downstream retrieval.

\subsection{Objective}
\label{sec:objective}
The objective of our framework is to balance token consumption and retrieval effectiveness in knowledge graph construction for RAG.
Unlike prior approaches that pursue maximum accuracy regardless of cost, we focus on reducing the total token consumption $T_{\text{total}}$ while retaining acceptable task performance.
Formally, we aim to solve the following trade-off problem:
\begin{equation}
\min T_{\text{total}} \quad \text{subject to} \quad \text{Accuracy}(RAG(G)) \geq \delta
\label{eq:objective1}
\end{equation}
where $\delta$ denotes a task-dependent performance threshold.
The concrete evaluation metrics used to instantiate $\text{Accuracy}(RAG(G))$ (e.g., Exact Match, F1) are described in the experimental section. 
Given that our primary goal is to reduce token consumption by one to two orders of magnitude, we set $\delta$ relative to a strong baseline, requiring our framework to achieve at least 80\% of the accuracy obtained by AutoSchemaKG combined with HippoRAG1 (Full-KG) on each dataset. This pragmatic threshold ensures that our method remains highly effective and competitive for practical applications.

\section{Pipeline} 
In this section, we provide a detailed description of our pipeline for constructing a Knowledge Graph from source documents.

\subsection{Named Entity and Document-Level Concept Extraction}

Inspired by MiniRAG~\cite{fan2025miniragextremelysimpleretrievalaugmented} and the recent survey of Wang et al.~\cite{peng2022copenprobingconceptualknowledge}, 
which proposes a unified taxonomy of conceptualization across multiple semantic levels 
(entity, event, document, and system), we deliberately restrict our extraction to only 
\textbf{named entities} and \textbf{document-level concepts}. 
This design choice simplifies KG construction and substantially reduces token consumption, 
while still preserving the essential semantic units required for effective retrieval.

\subsubsection{Named Entities}
We adopt the definition of Named Entity Recognition (NER) following Petasis et al.~\cite{10.1145/345508.345563}, where ``a Named Entity (NE) is a proper noun serving as a name for something or someone.'' Our goal is to extract canonical mentions of salient entities.

Formally, given a passage $p$, the NER model extracts a set of entity mentions:
\[
\mathcal{E}(p) = \{ e_1, e_2, \dots, e_m \}.
\]

All extracted entities across passages are aggregated into a unified concept node set:
\[
V_{\mathrm{ent}} = \bigcup_{p \in V_{\mathrm{pas}}} \mathcal{E}(p),
\]
where $V_{\mathrm{pas}}$ denotes the set of passage nodes in the knowledge graph.

Each entity $e \in V_{\mathrm{ent}}$ is treated as an atomic node in the KG.

\subsubsection{Document-Level Concepts}   
As defined by Wang et al.~\cite{peng2022copenprobingconceptualknowledge}, 
document-level conceptualization abstracts information at the passage or document scale, 
capturing the main ideas and context beyond individual entities or events.

\subsubsection{Prompt Design} 
To improve the effectiveness of both NER and concept extraction, 
we adopt a \textit{few-shot prompting} strategy rather than zero-shot instructions as it can significantly improve modern LLMs' performance ~\cite{brown2020languagemodelsfewshotlearners}. 
The LLM is provided with several annotated examples, which guide it to produce  
more accurate and consistent extractions. 
Furthermore, to minimize token consumption, the model is instructed to output 
only the extracted entities or concepts directly, instead of generating 
structured JSON. This design reduces output verbosity and significantly 
lowers the number of tokens required per query.

\subsubsection{Concept Deduplication}
Since duplicate evidences and concepts can reduce RAG accuracy~\cite{ko-etal-2024-evidence}, 
we apply a strict deduplication procedure that merges only nodes with identical \textit{type} and \textit{name}. 
This process yields a cleaner and more connected knowledge graph while minimizing the introduction of noisy nodes. 
Such strict merging is particularly important during the query phase, 
as named entities extracted from queries will be aligned with their corresponding concept nodes in the graph.

Formally, let $\mathcal{V}_{con} = \{(t_i, n_i)\}_{i=1}^N$ denote the set of extracted concept nodes, 
where $t_i$ and $n_i$ represent the \textit{type} and \textit{name} of the $i$-th concept. 
The deduplicated set of concept nodes $\mathcal{V}_{con}^\ast$ is defined as
\begin{equation}
\mathcal{V}_{con}^\ast = \{ (t, n) \mid \exists\, i \ \text{s.t.}\ (t, n) = (t_i, n_i) \},
\label{eq:objective2}
\end{equation}
which ensures that each unique pair of \textit{type} and \textit{name} appears only once in the knowledge graph.

\subsection{Graph Construction}
Based on the extracted entities and concepts, and cluster results we create a graph by adding three types of nodes mentioned in the problem definition.

\paragraph{Passage Linkage}
Each concept node is linked to the passage from which it was extracted. 
This preserves provenance and ensures that retrieval can always trace nodes back to their textual source. These edges are directed from $V_{\mathrm{con}}$ to $V_{\mathrm{pas}}$, as defined in Section~2.2.

\paragraph{Co-Occurrence Edges}
If two nodes appear in the same passage, we create a bidirectional 
\texttt{co\_occurrence} edge between them. This encodes local contextual 
associations between entities and concepts. 

\subsection{Retrieval}
In the retrieval phase, we adopt a lightweight design that minimizes LLM usage. 
The LLM is only applied for query-level NER and final answer generation, 
while the core retrieval relies entirely on non-LLM methods to reduce token consumption.

\subsubsection{Query NER}
Similar to entity and concept extraction in document processing,
we use a few-shot prompt to extract named entities from the query.
These extracted items are then matched against the concept node set $V_{\mathrm{con}}$
in the knowledge graph.
We first attempt exact string matching; if no matches are found,
we select the top-$3$ nodes with the highest semantic similarity
(based on embeddings).
The resulting matched nodes are used to construct a \textit{personalized dictionary}
that serves as the restart distribution for PPR.

\subsubsection{Personalized PageRank}
We run a PPR algorithm on the knowledge graph, 
biased toward the query-relevant nodes identified in the previous step. 
Each matched node $u$ is assigned a weight based on two factors: 
semantic relevance and frequency. 
For nodes matched by exact string matching, the semantic weight is set to $1$. 
For nodes matched by semantic similarity, the semantic weight is given by the 
similarity score $s(u)$ between the query and the node embedding. 
In both cases, the frequency weight is defined as the inverse of the 
node frequency $f(u)$ in the corpus. 
Formally, the unnormalized weight $w(u)$ is defined as
\begin{equation}
w(u) = \frac{s(u)}{f(u)},
\label{eq:weight_full}
\end{equation}

where
\begin{equation}
s(u) =
\begin{cases}
1, & \text{if } u \text{ is an exact match}, \\
\text{sim}(q, u), & \text{if } u \text{ is selected by semantic similarity}.
\end{cases}
\label{eq:s_definition}
\end{equation}

To avoid imbalance caused by differing numbers of exact and semantic matches, 
the weights are normalized within each group:
\begin{equation}
\hat{w}(u) = \frac{w(u)}{\sum_{v \in G} w(v)}, \quad u \in G
\label{eq:normalized_weight}
\end{equation}
where $G$ denotes either the exact-match group or the semantic-match group. 
The final personalized dictionary is constructed from $\hat{w}(u)$ across both groups, 
and serves as the teleportation vector for PPR.
After running PPR on the knowledge graph, we rank passages by their visiting frequencies 
and select the top 5 passages. These passages are then provided to the reader model for final answer generation.

\section{Experiment and Result}
This section describes the experimental setup and reports the results.
\subsection{Experimental Setup}
\subsubsection{Datasets.}
Following AutoSchemaKG, we evaluate our method on three benchmark multi-hop QA datasets: 
MuSiQue, HotpotQA and 2WikiMultihopQA~\cite{trivedi2022musiquemultihopquestionssinglehop,yang2018hotpotqadatasetdiverseexplainable,ho2020constructingmultihopqadataset}. 
These datasets are established benchmarks for multi-hop QA, each emphasizing distinct aspects such as connected reasoning (MuSiQue), explainability through supporting facts (HotpotQA), and structured evidence with reasoning paths (2WikiMultihopQA). 
Together, they provide a diverse and rigorous benchmark for evaluating multi-document reasoning.

\subsubsection{Models Used}
To ensure consistency with the data from AutoSchemaKG, we employ Meta's LLaMA-3.1-8B-Instruct for entity and concept extraction, and LLaMA-3.3-70B-Instruct for answer generation. These models exhibit strong reasoning and summarization capabilities, and being open-source, they are particularly suitable for our study~\cite{meta2024llama31}.

\subsubsection{Baseline and Metrics}
For retrieval accuracy, we compare our method against several representative RAG approaches, including LightRAG, MiniRAG,\\
GraphRAG, and AutoSchemaKG.
We select LightRAG and MiniRAG because they are designed as lightweight graph-based RAG methods. 
We include \\ GraphRAG as it is one of the most widely adopted graph-based RAG approaches, 
and AutoSchemaKG because it directly inspired our design.  
For efficiency, measured in terms of token consumption, we use 
LightRAG, MiniRAG, and AutoSchemaKG (with HippoRAG1 module) as references.

For evaluation, we report two standard metrics EM and F1, as well as token consumption:
\begin{itemize}
    \item \textbf{Exact Match (EM).} 
    EM measures the proportion of predictions that exactly match the 
    ground-truth answer string after standard normalization 
    (e.g., lowercasing and punctuation removal). Formally, if 
    $y_i$ denotes the predicted answer for the $i$-th question and 
    $y_i^*$ its ground truth, EM is defined as
    \begin{equation}
    \text{EM} = \frac{1}{N} \sum_{i=1}^{N} \mathbf{1}[\,y_i = y_i^*\,]
    \label{eq:em_metric}
    \end{equation}

    where $\mathbf{1}[\cdot]$ is the indicator function and $N$ is the number of questions.

    \item \textbf{F1 score.} 
    F1 measures the token-level overlap between predictions and ground-truth 
    answers, capturing both precision and recall. Let $P_i$ and $G_i$ denote 
    the sets of tokens in the predicted and ground-truth answers for the 
    $i$-th question. Precision and recall are defined as
    \begin{align}
    \text{Precision}_i &= \frac{|P_i \cap G_i|}{|P_i|}, \label{eq:precision} \\
    \text{Recall}_i    &= \frac{|P_i \cap G_i|}{|G_i|}. \label{eq:recall}
    \end{align}
    The F1 score for the $i$-th instance is then
    \begin{equation}
    \text{F1}_i = \frac{2 \cdot \text{Precision}_i \cdot \text{Recall}_i}
                        {\text{Precision}_i + \text{Recall}_i}
    \label{eq:f1_metric}
    \end{equation}
    and the overall F1 is obtained by averaging across all $N$ questions.

    \item \textbf{Token Consumption.} 
    As defined in Section~2, we report input, prompt, and output tokens 
    as our efficiency metric.
\end{itemize}

\subsection{Retrieval Results}
\vspace{-2pt}
\begin{table*}[!htbp]
\scriptsize
\centering
\caption{Retrieval accuracy (EM/F1) and relative output token usage of LightRAG, MiniRAG, and TERAG on MuSiQue, 2Wiki, and HotpotQA datasets.}
\label{tab:baseline_light_mini}
\begin{tabular}{lccccccccc}
\toprule
\multirow{2}{*}{Model} & 
\multicolumn{3}{c}{MuSiQue} & 
\multicolumn{3}{c}{2Wiki} & 
\multicolumn{3}{c}{HotpotQA} \\
\cmidrule(lr){2-4}\cmidrule(lr){5-7}\cmidrule(lr){8-10}
& EM & F1 & Rel.~(\%) & EM & F1 & Rel.~(\%) & EM & F1 & Rel.~(\%) \\
\midrule
LightRAG       & \textbf{20.0} & 29.3 & 2753 & 38.6 & 44.6 & 1434 & 33.3 & 44.9 & 1041 \\
MiniRAG        &  9.6 & 16.8 & 4145 & 13.2 & 21.4 & 1602 & \textbf{47.1} & 59.8 & 2553 \\
TERAG          & 18.8 & \textbf{29.6} & \textbf{100} & \textbf{51.2} & \textbf{57.8} & \textbf{100} & 46.9 & \textbf{59.8} & \textbf{100} \\
\bottomrule
\end{tabular}
\end{table*}

\begin{table}[!htbp]
\centering
\scriptsize
\caption{Retrieval accuracy (EM/F1) on MuSiQue, 2Wiki, and HotpotQA datasets. Results for all baseline RAG methods are taken from \cite{bai2025autoschemakgautonomousknowledgegraph}, while TERAG is reported from our own experiments. Best results in each column are highlighted in bold.}
\resizebox{\linewidth}{!}{%
\begin{tabular}{lcccccc}
\toprule
\multirow{2}{*}{Model} &
\multicolumn{2}{c}{MuSiQue} &
\multicolumn{2}{c}{2Wiki} &
\multicolumn{2}{c}{HotpotQA} \\
\cmidrule(lr){2-3}\cmidrule(lr){4-5}\cmidrule(lr){6-7}
& EM & F1 & EM & F1 & EM & F1 \\
\midrule
\multicolumn{7}{l}{\textbf{Baseline Retrievers}} \\
No Retriever      & 17.6 & 26.1 & 36.5 & 42.8 & 37.0 & 47.3 \\
Contriever        & 24.0 & 31.3 & 38.1 & 41.9 & 51.3 & 62.3 \\
BM25              & 20.3 & 28.8 & 47.9 & 51.2 & 52.0 & 63.4 \\
\midrule
\multicolumn{7}{l}{\textbf{Existing Graph-based RAG Methods}} \\
GraphRAG                 & \textbf{27.3} & \textbf{38.5} & 51.4 & 58.6 & \textbf{55.2} & \textbf{68.6} \\
LightRAG                 & 20.0 & 29.3 & 38.6 & 44.6 & 33.3 & 44.9 \\
MiniRAG                  &  9.6 & 16.8 & 13.2 & 21.4 & 47.1 & 59.9 \\
AutoSchemaKG + HippoRAG1 & 23.6 & 36.5 & \textbf{54.8} & \textbf{63.2} & 50.0 & 65.3 \\
\midrule
TERAG (Ours)             & 18.8 & 29.6 & 51.2 & 57.8 & 46.9 & 59.8 \\
\midrule
Target (80\% of AutoSchemaKG + HippoRAG1) 
                         & 18.8 & 29.2 & 43.9 & 50.6 & 40.0 & 52.2 \\
\bottomrule
\end{tabular}}
\label{tab:retrieval-results}
\end{table}

The retrieval accuracy compared with other lightweight graph-based RAG methods on the three datasets is summarized in Table~\ref{tab:baseline_light_mini}.
Using LLaMA-3 8B as the graph construction model and LLaMA-3 70B as the reader model,
our token-efficient graph framework outperforms two popular lightweight graph-based RAG systems, LightRAG and MiniRAG, on most tasks while consuming only 3–10\% of their tokens.
Compared with AutoSchemaKG, our framework also meets the predefined performance target from Section~\ref{sec:objective}, which requires achieving at least 80\% of the accuracy of AutoSchemaKG + HippoRAG1 (Full-KG) on each dataset.
The complete accuracy comparison is provided in Table~\ref{tab:retrieval-results}.
Notably, on the 2Wiki dataset our method achieves accuracy close to the widely used GraphRAG
(EM: 51.2 vs.\ 51.4; F1: 57.8 vs.\ 58.6) while consuming substantially fewer tokens.
\vspace{-2pt}
\subsection{Token Consumption}
\vspace{2pt}
\begin{table}[t]
\caption{Token consumption statistics of our method (TERAG), LightRAG, MiniRAG and AutoSchemaKG across datasets.}
\label{tab:our_token_usage}
\centering

\footnotesize  
\renewcommand{\arraystretch}{0.9} 
\setlength{\aboverulesep}{2pt} 
\setlength{\belowrulesep}{2pt} 

\begin{tabular*}{\linewidth}{@{\extracolsep{\fill}} l
  >{\raggedright\arraybackslash}p{0.23\linewidth}
  S[table-format=8.0]
  S[table-format=8.0]
  S[table-format=8.0]}
\toprule
Method & Dataset & {Input} & {Output} & {Total} \\
\midrule
\textbf{TERAG (ours)} & HotpotQA
  & {\bfseries \num{2005645}} & {\bfseries \num{562827}} & {\bfseries \num{2568472}} \\
\textbf{TERAG (ours)} & \makecell[tl]{2WikiMultihopQA}
  & {\bfseries \num{1211644}} & {\bfseries \num{368708}} & {\bfseries \num{1580352}} \\
\textbf{TERAG (ours)} & MuSiQue
  & {\bfseries \num{2355941}} & {\bfseries \num{664702}} & {\bfseries \num{3020643}} \\
\midrule
AutoSchemaKG & HotpotQA        & \num{5723733}  & \num{4915796}  & \num{10639529} \\
AutoSchemaKG & \makecell[tl]{2WikiMultihopQA} & \num{3596676}  & \num{3176095}  & \num{6772771}  \\
AutoSchemaKG & MuSiQue         & \num{8960502}  & \num{7715976}  & \num{16676478} \\
\midrule
LightRAG     & HotpotQA        & \num{38765230} & \num{5862363}  & \num{44627593} \\
LightRAG     & \makecell[tl]{2WikiMultihopQA} & \num{34222643} & \num{5288806}  & \num{39511449} \\
LightRAG     & MuSiQue         & \num{68500000} & \num{18300000} & \num{86800000} \\
\midrule
MiniRAG      & HotpotQA        & \num{36909150} & \num{14370416} & \num{51279566} \\
MiniRAG      & \makecell[tl]{2WikiMultihopQA} & \num{13877889} & \num{5906984}  & \num{19784873} \\
MiniRAG      & MuSiQue         & \num{62425404} & \num{27552031} & \num{89977435} \\
\cmidrule(r){1-5}
\multicolumn{5}{@{}p{\linewidth}@{}}{\textit{Note: Some data from the HippoRAG2 paper~\cite{gutiérrez2025ragmemorynonparametriccontinual}.}} \\
\bottomrule
\end{tabular*}
\end{table}

While the retrieval accuracy of TERAG is comparable to lightweight graph-based RAG baselines, our key advantage is token efficiency.
Table~\ref{tab:our_token_usage} summarizes end-to-end token usage across datasets.
On HotpotQA, 2WikiMultihopQA, and MuSiQue, AutoSchemaKG consumes \textbf{8.6–11.6}× more completion (output) tokens and \textbf{2.9–3.8}× more input tokens than TERAG.
This advantage becomes even more pronounced against LightRAG and MiniRAG.LightRAG consumes \textbf{19–29}× more input tokens and \textbf{10–28}× more output tokens, while MiniRAG uses \textbf{11–27}× more input tokens and \textbf{16–42}× more output tokens. This highlights that our method spends \textbf{88-97\%} less token compared with other lightweight graph RAG methods.
(see Table~\ref{tab:our_token_usage_relative} for percentages normalized to TERAG=100). 
\begin{table}[ht]
\scriptsize
\caption{Relative token consumption across datasets. Manually Set TERAG = 100\%.}
\label{tab:our_token_usage_relative}
\begin{tabular*}{\linewidth}{@{\extracolsep{\fill}} ll
S[table-format=4.1]
S[table-format=4.1]
S[table-format=4.1]}
\toprule
Method & Dataset & {Input (\%)} & {Output (\%)} & {Total (\%)} \\
\midrule
\textbf{TERAG (ours)} & HotpotQA & 100.0 & 100.0 & 100.0 \\
& 2WikiMultihopQA & 100.0 & 100.0 & 100.0 \\
& MuSiQue & 100.0 & 100.0 & 100.0 \\
\midrule
AutoSchemaKG & HotpotQA & 285.4 & 873.4 & 414.2 \\
& 2WikiMultihopQA & 296.8 & 861.4 & 428.6 \\
& MuSiQue & 380.3 & 1160.8 & 552.1 \\
\midrule
LightRAG & HotpotQA & 1940.0 & 1041.6 & 1737.5 \\
& 2WikiMultihopQA & 2824.5 & 1434.4 & 2500.2 \\
& MuSiQue & 2907.5 & 2753.1 & 2873.5 \\
\midrule
MiniRAG & HotpotQA & 1840.3 & 2553.3 & 1996.5 \\
& 2WikiMultihopQA & 1145.4 & 1602.1 & 1251.9 \\
& MuSiQue & 2649.7 & 4145.0 & 2978.8 \\
\bottomrule
\end{tabular*}
\end{table}
This saving is practically important because LLM inference is dominated by the \emph{autoregressive decoding} stage: output tokens are generated one-by-one, and every step must read the accumulated KV cache and append new K/V pairs, which raises per-output-token latency and memory-bandwidth cost; by contrast, input tokens are processed in a parallel prefill pass \cite{zhou2024surveyefficientinferencelarge}.
Architecturally, our pipeline is token-efficient because we \emph{directly} extract graph concepts from passages in a single pass, avoiding the multi-stage LLM extraction/summarization used by prior graph-RAG systems such as AutoSchemaKG and LightRAG \cite{bai2025autoschemakgautonomousknowledgegraph,guo2025lightragsimplefastretrievalaugmented}.

\begin{figure}[ht]
    \centering
    \caption{Accuracy versus output token consumption on the 2Wiki dataset. The upper-right region indicates higher accuracy with lower token consumption. }
    \includegraphics[width=0.9\linewidth]{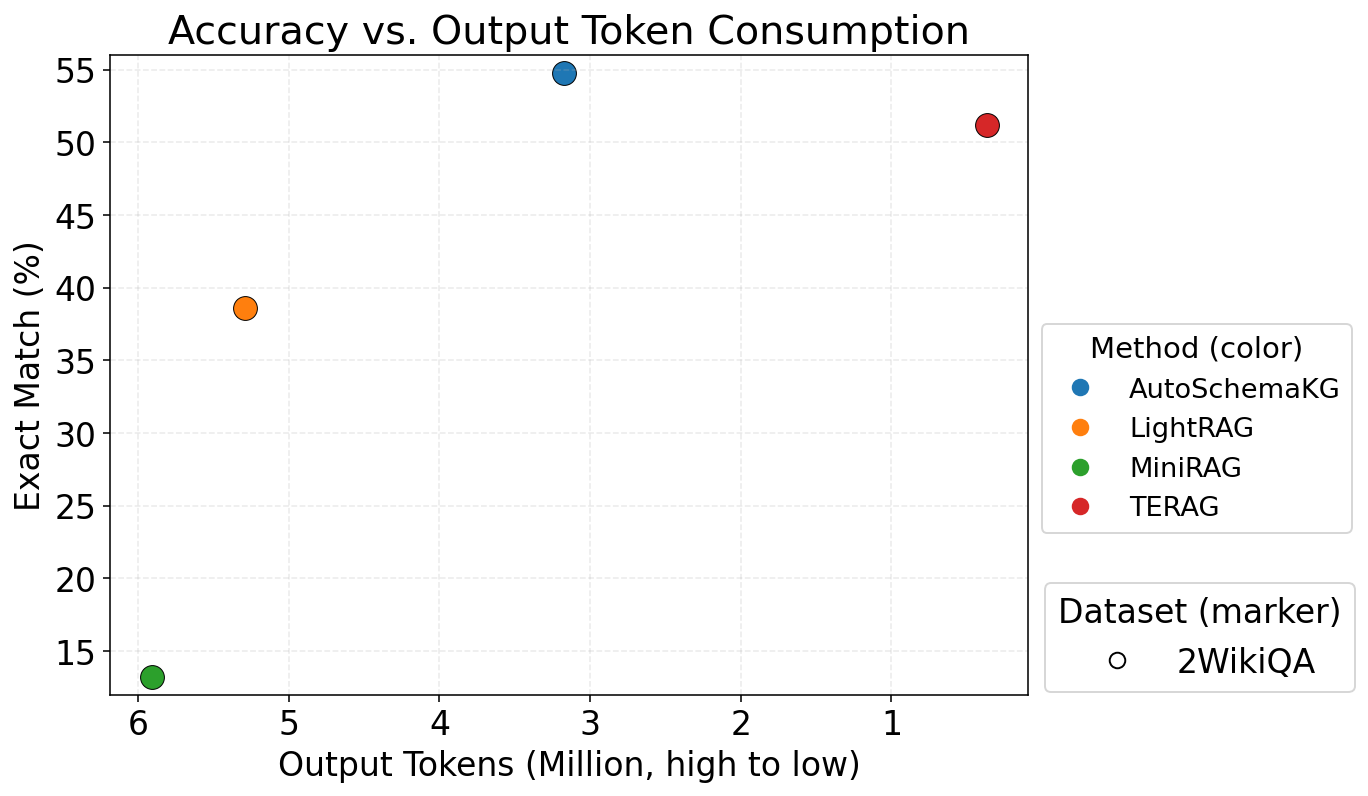}
    \label{fig:em_vs_token}
\end{figure}

Figure~\ref{fig:em_vs_token} provides a clear visualization of the relationship between output token consumption and retrieval accuracy across different graph-based RAG methods. The x-axis represents the number of output tokens generated by each model, while the y-axis reports the EM score, allowing us to directly examine the efficiency–effectiveness trade-off. An ideal token-efficient method would appear toward the upper-right corner, combining high retrieval accuracy with low token usage. Our proposed TERAG consistently lies closest to this desirable region, indicating that it achieves competitive accuracy while consuming substantially fewer tokens. In contrast, AutoSchemaKG attains strong accuracy but only by incurring an order-of-magnitude increase in token usage, making it far less cost-efficient. LightRAG, on the other hand, reduces token usage but at the expense of a sharp drop in accuracy, failing to provide a balanced trade-off. Taken together, these results confirm that TERAG delivers the most favorable efficiency–performance balance among the evaluated methods, highlighting its practicality for large-scale or resource-constrained deployment scenarios.

\FloatBarrier

\section{Ablation Study}

\paragraph{Effect of Restart Vector Design.}
We examine the impact of our new PPR restart vector formulation, which integrates both semantic relevance and concept frequency. Unlike traditional PPR methods that rely solely on uniform or inverse-frequency distributions, our approach aims to produce a more informative restart vector that better captures contextual importance.

\begin{table}[ht]
\centering
\scriptsize
\caption{Ablation study on PPR restart vector calculation. ``New`` denotes our method combining semantic relevance and frequency, while ``Original`` uses only inverse frequency.}
\label{tab:ablation_results}
\resizebox{\linewidth}{!}{%
\begin{tabular}{l S[table-format=2.1, table-space-text-post={(+5\%)}]
                   S[table-format=2.1, table-space-text-post={(+5\%)}]
                   S[table-format=2.1, table-space-text-post={(+5\%)}]
                   S[table-format=2.1, table-space-text-post={(+5\%)}]
                   S[table-format=2.1, table-space-text-post={(+5\%)}]
                   S[table-format=2.1, table-space-text-post={(+5\%)}]}
\toprule
\multirow{2}{*}{Model/Dataset} &
\multicolumn{2}{c}{MuSiQue} &
\multicolumn{2}{c}{2Wiki} &
\multicolumn{2}{c}{HotpotQA} \\
\cmidrule(lr){2-3}\cmidrule(lr){4-5}\cmidrule(lr){6-7}
& {EM} & {F1} & {EM} & {F1} & {EM} & {F1} \\
\midrule
TERAG + New & 18.7 {(+3\%)} & 29.6 {(+2\%)} & 51.2 {(+3\%)} & 57.8 {(+3\%)} & 46.9 {(+5\%)} & 59.8 {(+2\%)} \\
TERAG + Original & 18.3 & 29.1 & 49.7 & 56.1 & 44.8 & 58.7 \\
\bottomrule
\end{tabular}
}
\end{table}

As shown in Table~\ref{tab:ablation_results}, our restart vector design consistently improves retrieval accuracy across all three datasets, yielding gains of 2--5 percentage points in both EM and F1. These results highlight the importance of combining semantic and frequency information to construct a more discriminative restart distribution, leading to more precise passage retrieval.

\paragraph{Model Variation Analysis.}
We also assessed the influence of switching the answer generation model while keeping the same retrieval graph. Replacing the default model with \texttt{LLaMA-3.2 11B} caused a slight performance drop, while \texttt{Qwen2 235B} achieved accuracy similar to the \texttt{LLaMA-3 70B} but introduced noticeably higher latency. These observations suggest that the dominant performance factor in our framework is the model’s NER accuracy; improvements in generic reasoning or generation capabilities do not yield proportional gains in retrieval quality.

\FloatBarrier
\section{Conclusion and Future Work}
To address the high token consumption incurred by current Graph-RAG systems during knowledge graph construction, we propose the TERAG framework. TERAG constructs graph structures by directly extracting named entities and document-level concepts from text in a single pass, thereby eliminating the multiple and costly LLM reasoning calls required by prior approaches. 

Despite its lightweight design, TERAG achieves highly competitive retrieval accuracy on multiple multi-hop QA benchmarks while reducing token overhead by one to two orders of magnitude. In addition, because concept extraction, concept filtering, and graph linking are largely independent, these components can be parallelized efficiently. As a result, TERAG not only minimizes indexing-related token cost but also has the potential to significantly shorten graph construction latency in large-scale deployments. Overall, our findings demonstrate that a carefully designed lightweight graph construction pipeline can achieve a more favorable efficiency–performance trade-off than graph-RAG systems that heavily rely on LLMs.

For future work, we plan to explore extracting concepts with controllable levels of granularity, enabling multi-layered knowledge graphs that better capture semantic structure and improve retrieval accuracy. We also aim to incorporate additional engineering strategies to further optimize the balance between token efficiency and model effectiveness.

Finally, due to TERAG’s simple and adaptable pipeline, we expect its performance to remain stable under evolving data streams or more heterogeneous corpora. However, the lack of widely recognized benchmarks for such dynamic and heterogeneous retrieval scenarios limits empirical evaluation. Future research will involve designing appropriate datasets and evaluation protocols to more systematically validate these capabilities.

%
%
\FloatBarrier
\bibliographystyle{splncs04} 
\bibliography{TERAG} 

\begin{thebibliography}{10}
\providecommand{\url}[1]{\texttt{#1}}
\providecommand{\urlprefix}{URL }
\providecommand{\doi}[1]{https://doi.org/#1}

\bibitem{meta2024llama31}
AI, M.: Introducing llama 3.1: Our most capable models to date (July 23 2024), \url{https://ai.meta.com/blog/meta-llama-3-1/}, accessed: YYYY-MM-DD

\bibitem{bai2025autoschemakgautonomousknowledgegraph}
Bai, J., Fan, W., Hu, Q., Zong, Q., Li, C., Tsang, H.T., Luo, H., Yim, Y., Huang, H., Zhou, X., Qin, F., Zheng, T., Peng, X., Yao, X., Yang, H., Wu, L., Ji, Y., Zhang, G., Chen, R., Song, Y.: Autoschemakg: Autonomous knowledge graph construction through dynamic schema induction from web-scale corpora (2025), \url{https://arxiv.org/abs/2505.23628}

\bibitem{brown2020languagemodelsfewshotlearners}
Brown, T.B., Mann, B., Ryder, N., Subbiah, M., Kaplan, J., Dhariwal, P., Neelakantan, A., Shyam, P., Sastry, G., Askell, A., Agarwal, S., Herbert-Voss, A., Krueger, G., Henighan, T., Child, R., Ramesh, A., Ziegler, D.M., Wu, J., Winter, C., Hesse, C., Chen, M., Sigler, E., Litwin, M., Gray, S., Chess, B., Clark, J., Berner, C., McCandlish, S., Radford, A., Sutskever, I., Amodei, D.: Language models are few-shot learners (2020), \url{https://arxiv.org/abs/2005.14165}

\bibitem{xue2023dbgpt}
Deng, X., Wang, B., Chen, J., Gan, Z., Liu, J., Shi, W., Wang, Y., Wang, F., Zhang, J., Zhang, X.: {DB-GPT}: Empowering database interactions with private large language models (2023)

\bibitem{Edge2024FromLT}
Edge, D., Trinh, H., Cheng, N., Bradley, J., Chao, A., Mody, A., Truitt, S., Larson, J.: From local to global: A graph rag approach to query-focused summarization. ArXiv  \textbf{abs/2404.16130} (2024), \url{https://api.semanticscholar.org/CorpusID:269363075}

\bibitem{edge2024lazygraphrag}
Edge, D., Trinh, H., Larson, J.: Lazygraphrag: Setting a new standard for quality and cost. \url{https://www.microsoft.com/en-us/research/blog/lazygraphrag-setting-a-new-standard-for-quality-and-cost/} (Nov 2024), microsoft Research Blog

\bibitem{fan2025miniragextremelysimpleretrievalaugmented}
Fan, T., Wang, J., Ren, X., Huang, C.: Minirag: Towards extremely simple retrieval-augmented generation (2025), \url{https://arxiv.org/abs/2501.06713}

\bibitem{fan2024goldcoingroundinglargelanguage}
Fan, W., Li, H., Deng, Z., Wang, W., Song, Y.: Goldcoin: Grounding large language models in privacy laws via contextual integrity theory (2024), \url{https://arxiv.org/abs/2406.11149}

\bibitem{ghimire2024generativeaieducationstudy}
Ghimire, A., Prather, J., Edwards, J.: Generative ai in education: A study of educators' awareness, sentiments, and influencing factors (2024), \url{https://arxiv.org/abs/2403.15586}

\bibitem{guo2025lightragsimplefastretrievalaugmented}
Guo, Z., Xia, L., Yu, Y., Ao, T., Huang, C.: Lightrag: Simple and fast retrieval-augmented generation (2025), \url{https://arxiv.org/abs/2410.05779}

\bibitem{gutierrez2024hipporag}
Gutierrez, B.J., Shu, Y., Gu, Y., Yasunaga, M., Su, Y.: Hippo{RAG}: Neurobiologically inspired long-term memory for large language models. In: The Thirty-eighth Annual Conference on Neural Information Processing Systems (2024), \url{https://openreview.net/forum?id=hkujvAPVsg}

\bibitem{gutiérrez2025ragmemorynonparametriccontinual}
Gutiérrez, B.J., Shu, Y., Qi, W., Zhou, S., Su, Y.: From rag to memory: Non-parametric continual learning for large language models (2025), \url{https://arxiv.org/abs/2502.14802}

\bibitem{ho2020constructingmultihopqadataset}
Ho, X., Nguyen, A.K.D., Sugawara, S., Aizawa, A.: Constructing a multi-hop qa dataset for comprehensive evaluation of reasoning steps (2020), \url{https://arxiv.org/abs/2011.01060}

\bibitem{hu-etal-2025-grag}
Hu, Y., Lei, Z., Zhang, Z., Pan, B., Ling, C., Zhao, L.: Grag: Graph retrieval-augmented generation. In: Findings of the Association for Computational Linguistics: NAACL 2025. pp. 4145--4157. Association for Computational Linguistics, Albuquerque, New Mexico (Apr 2025). \doi{10.18653/v1/2025.findings-naacl.232}, \url{https://aclanthology.org/2025.findings-naacl.232/}

\bibitem{huang2025ketragcostefficientmultigranularindexing}
Huang, Y., Zhang, S., Xiao, X.: Ket-rag: A cost-efficient multi-granular indexing framework for graph-rag (2025), \url{https://arxiv.org/abs/2502.09304}

\bibitem{ko-etal-2024-evidence}
Ko, S., Cho, H., Chae, H., Yeo, J., Lee, D.: Evidence-focused fact summarization for knowledge-augmented zero-shot question answering. In: Al-Onaizan, Y., Bansal, M., Chen, Y.N. (eds.) Proceedings of the 2024 Conference on Empirical Methods in Natural Language Processing. pp. 10636--10651. Association for Computational Linguistics, Miami, Florida, USA (Nov 2024). \doi{10.18653/v1/2024.emnlp-main.594}, \url{https://aclanthology.org/2024.emnlp-main.594/}

\bibitem{lewis2021retrievalaugmentedgenerationknowledgeintensivenlp}
Lewis, P., Perez, E., Piktus, A., Petroni, F., Karpukhin, V., Goyal, N., Küttler, H., Lewis, M., tau Yih, W., Rocktäschel, T., Riedel, S., Kiela, D.: Retrieval-augmented generation for knowledge-intensive nlp tasks (2021), \url{https://arxiv.org/abs/2005.11401}

\bibitem{liu2024surveymedicallargelanguage}
Liu, L., Yang, X., Lei, J., Shen, Y., Wang, J., Wei, P., Chu, Z., Qin, Z., Ren, K.: A survey on medical large language models: Technology, application, trustworthiness, and future directions (2024), \url{https://arxiv.org/abs/2406.03712}

\bibitem{marcus2020decadeaistepsrobust}
Marcus, G.: The next decade in ai: Four steps towards robust artificial intelligence (2020), \url{https://arxiv.org/abs/2002.06177}

\bibitem{graphrag_github_misc_2025}
{Microsoft}: {GraphRAG: Unlocking the Power of Private Data with LLM-Powered Graph RAG}. \url{https://github.com/microsoft/graphrag} (2025), accessed: 2025-09-06

\bibitem{NebulaGraphGraphRAG}
{NebulaGraph}: Graph rag: Unleashing the power of knowledge graphs with llm (September 2023), \url{https://www.nebula-graph.io/posts/graph-RAG}

\bibitem{peng2024graphretrievalaugmentedgenerationsurvey}
Peng, B., Zhu, Y., Liu, Y., Bo, X., Shi, H., Hong, C., Zhang, Y., Tang, S.: Graph retrieval-augmented generation: A survey (2024), \url{https://arxiv.org/abs/2408.08921}

\bibitem{peng2022copenprobingconceptualknowledge}
Peng, H., Wang, X., Hu, S., Jin, H., Hou, L., Li, J., Liu, Z., Liu, Q.: Copen: Probing conceptual knowledge in pre-trained language models (2022), \url{https://arxiv.org/abs/2211.04079}

\bibitem{10.1145/345508.345563}
Petasis, G., Cucchiarelli, A., Velardi, P., Paliouras, G., Karkaletsis, V., Spyropoulos, C.D.: Automatic adaptation of proper noun dictionaries through cooperation of machine learning and probabilistic methods. In: Proceedings of the 23rd Annual International ACM SIGIR Conference on Research and Development in Information Retrieval. pp. 128--135. SIGIR '00, Association for Computing Machinery, New York, NY, USA (2000). \doi{10.1145/345508.345563}, \url{https://doi.org/10.1145/345508.345563}

\bibitem{petroni-etal-2019-language}
Petroni, F., Rockt{\"a}schel, T., Riedel, S., Lewis, P., Bakhtin, A., Wu, Y., Miller, A.: Language models as knowledge bases? In: Inui, K., Jiang, J., Ng, V., Wan, X. (eds.) Proceedings of the 2019 Conference on Empirical Methods in Natural Language Processing and the 9th International Joint Conference on Natural Language Processing (EMNLP-IJCNLP). pp. 2463--2473. Association for Computational Linguistics, Hong Kong, China (Nov 2019). \doi{10.18653/v1/D19-1250}, \url{https://aclanthology.org/D19-1250/}

\bibitem{sun2024thinkongraphdeepresponsiblereasoning}
Sun, J., Xu, C., Tang, L., Wang, S., Lin, C., Gong, Y., Ni, L.M., Shum, H.Y., Guo, J.: Think-on-graph: Deep and responsible reasoning of large language model on knowledge graph (2024), \url{https://arxiv.org/abs/2307.07697}

\bibitem{trivedi2022musiquemultihopquestionssinglehop}
Trivedi, H., Balasubramanian, N., Khot, T., Sabharwal, A.: Musique: Multihop questions via single-hop question composition (2022), \url{https://arxiv.org/abs/2108.00573}

\bibitem{yang2018hotpotqadatasetdiverseexplainable}
Yang, Z., Qi, P., Zhang, S., Bengio, Y., Cohen, W.W., Salakhutdinov, R., Manning, C.D.: Hotpotqa: A dataset for diverse, explainable multi-hop question answering (2018), \url{https://arxiv.org/abs/1809.09600}

\bibitem{zhou2024surveyefficientinferencelarge}
Zhou, Z., Ning, X., Hong, K., Fu, T., Xu, J., Li, S., Lou, Y., Wang, L., Yuan, Z., Li, X., Yan, S., Dai, G., Zhang, X.P., Dong, Y., Wang, Y.: A survey on efficient inference for large language models (2024), \url{https://arxiv.org/abs/2404.14294}

\end{thebibliography}

\end{document}